\documentclass[journal,twoside,web]{ieeecolor}
\usepackage{tmi}
\usepackage{cite}
\usepackage{amsmath,amssymb,amsfonts}
\usepackage{algorithmic}
\usepackage{graphicx}
\usepackage{textcomp}
\usepackage{epstopdf}
\begin{document}
\title{Reciprocal Landmark Detection and Tracking with Extremely Few Annotations}
\author{Jianzhe Lin, \IEEEmembership{Student Member, IEEE}, Ghazal Sahebzamani, \IEEEmembership{Student Member, IEEE}, Christina Luong, Fatemeh Taheri Dezaki, \IEEEmembership{Student Member, IEEE},Mohammad Jafari, \IEEEmembership{Student Member, IEEE}, Purang Abolmaesumi, \IEEEmembership{Fellow, IEEE}, and Teresa Tsang
\thanks{Jianzhe Lin, Ghazal Sahebzamani, Fatemeh Taheri Dezaki, Mohammad Jafari, and Purang Abolmaesumi are with the Electrical and Computer Engineering Department, University of British Columbia, Vancouver, BC, V6T 1Z2, Canada.(e-mail: jianzhelin, ghazal, fatemeht, mohammadj, purang@ece.ubc.ca).}
\thanks{Teresa Tsang is the Associate Head Research and Co-Acting Head, Department of Medicine,University of British Columbia. She is Director of Echo Laboratory at Vancouver General Hospital, Vancouver, BC, Canada.(e-mail:t.tsang@ubc.ca.}
\thanks{Christina Luong isa Clinical Assistant Professor within the Division of Cardiology at the University of British Columbia. She is the Head of Stress Echocardiography at Vancouver General Hospital,  Vancouver, BC, Canada..(e-mail:t.tsang@ubc.ca).}
}

\maketitle

\begin{abstract}
Localization of anatomical landmarks to perform two-dimensional measurements in echocardiography is part of routine clinical workflow in cardiac disease diagnosis. Automatic localization of those landmarks is highly desirable to improve workflow and reduce interobserver variability. Training a machine learning framework to perform such localization is hindered given the sparse nature of gold standard labels; only few percent of cardiac cine series frames are normally manually labeled for clinical use. In this paper, we propose a new end-to-end reciprocal detection and tracking model that is specifically designed to handle the sparse nature of echocardiography labels. The model is trained using few annotated frames across the entire cardiac cine sequence to generate consistent detection and tracking of landmarks, and an adversarial training for the model is proposed to take advantage of these annotated frames. The superiority of the proposed reciprocal model is demonstrated using a series of experiments.
\end{abstract}


\section{Introduction}
\begin{figure}
\centering
\includegraphics[width=\linewidth]{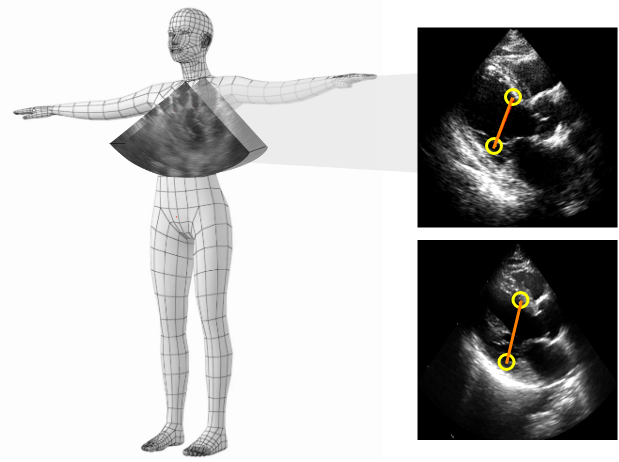}
\caption{Example of PLAX images, as one of the most common standard views acquired in point-of-care echocardiography. Landmarks identified on the left ventricle are used to measure the EF, a strong indicator of cardiac disease. Two landmarks on inferolateral and anteroseptal walls (IW, AW) are yellow color while the LVID is the red line. LVID can be localized with IW and AW.}
\label{fig:0}
\end{figure}

\IEEEPARstart{D}{ata} scarcity and lack of annotation is a general problem for developing machine learning models in medical imaging. Among various medical imaging modalities, ultrasound (US) is the most frequently used modality given its widespread availability, lower cost, and safety since it does not involve ionizing radiation. Specifically, US imaging, in the form of echocardiography (echo), is the standard-of-care in cardiac imaging for the detection of heart disease. Echo examinations are performed across up to 14 standard views from several acoustic windows on the chest. In this paper, we specifically focus on the parasternal long axis (PLAX), which is one of the most common view acquired in the point-of-care US for rapid examination of cardiac function (Fig.~\ref{fig:0}). Several measurements from PLAX require the localization of anatomical landmarks across discrete points in the cardiac cycle. Our work specifically investigates automatic localization of the left ventricle (LV) internal dimension (LVID), which is routinely used to estimate the ejection fraction (EF), a strong indicator of cardiac function abnormality. In clinics, LVID landmarks are determined in two frames of the cardiac cycle, i.e. end-diastolic and end-systolic. However, such annotation is challenging, specially for general physicians at point-of-care who do not have the experience of cardiologists. As such, the automation of landmark localization is highly desirable. However, developing a machine learning model for such automation has been hindered by the availability of only sparse set of labeled frames in cardiac cines. Manually labeling all cardiac frames for a large set of cardiac cines is virtually impractical, given limited expert time.


Instead of manually labeling, we propose a new Reciprocal landmark Detection and Tracking (RDT) model that enables automation in measurements across the entire cardiac cycle. The model only uses prior knowledge from sparsely labeled key frames that are temporally distant in a cardiac cycle. Meanwhile, we take advantage of temporal coherence of cardiac cine series to impose cycle consistency in tracking landmarks across unannotated frames that are between these two annotated frames. To impose consistent detection and tracking of the landmarks, we propose a reciprocal training as a self-supervision process.
\begin{figure*}[!ht]
\centering
\includegraphics[width=\linewidth]{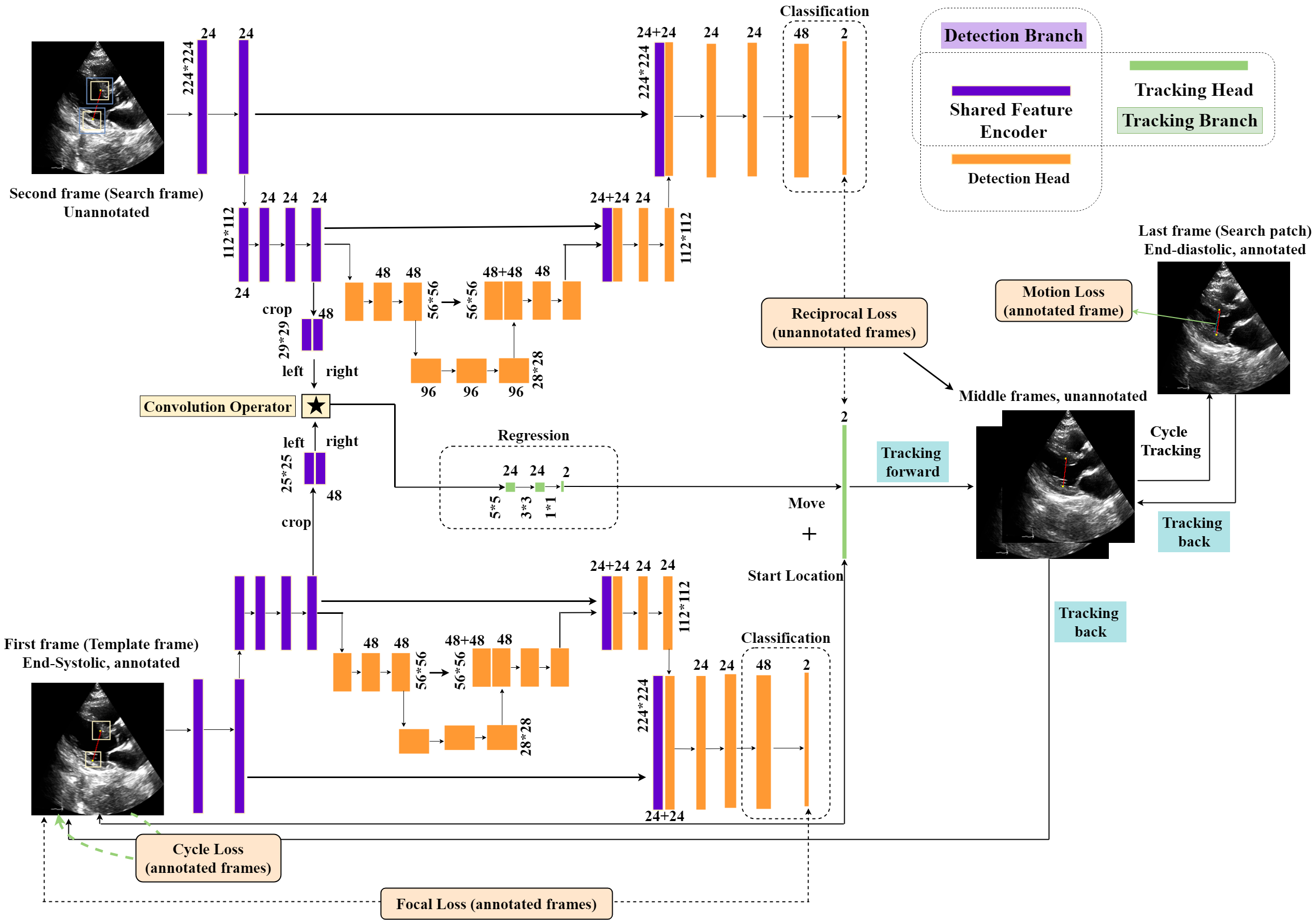}
\caption{The general flowchart of the proposed detection and tracking model. Gold standard labels are only available for end-diastolic and end-systolic frames. The propagation starts from the end-diastolic frame and ends at the end-systolic frame. The tracking is completed in a cycle way. The two annotated frames serve as a weak supervision for the model. The detection and tracking results from the unannotated frames jointly reciprocally provide another self-supervision.}
\label{fig:1}
\end{figure*}

In summary, we propose a RDT model, which is weakly supervised by only two annotated keyframes in an image sequence for model training. For testing, the model is an end-to-end model that detects the landmark in the first frame, followed by a tracking process. Our contributions are:
\begin{itemize}
    \item A novel Reciprocal landmark Detection and Tracking (RDT) model is proposed. In the model, the spatial constraint for detection and temporal coherence for tracking of cardiac cine series work reciprocally, to generate accurate localization of landmarks;
    \item The sparse nature of echocardiography labels is handled by the proposed model. The model is only weakly supervised by two annotated image frames that are temporally distant from each other. The annotation sparsity is also analyzed in the experimental part;
    \item A novel adversarial training approach (Ad-T) for optimization of the proposed RDT. Such training is made possible by introducing four complementary losses as in Fig.~\ref{fig:1}, i.e. reciprocal loss, motion loss, focal loss, and cycle loss. Compared with conventional training approaches, Ad-T indirectly achieves feature augmentation, which is extremely important for model training given the extremely few annotations. the advantage of such Ad-T is highlighted in our ablation study.
\end{itemize}

\section{Related Work}
As a low cost, low risk, and easily accessible modality, the cardiac US is widely used as an assessment tool in point-of-care. With the utilization of US technology in various form factors from cart-based to hand-held devices, measurement of cardiac structures can be typically conducted by users with diverse levels of expertise. However, due to US images' noisy nature, studies indicate large amounts of inter- and intra-observer variability even among experts \cite{thorstensen2010reproducibility}. This amount of observer variability may easily lead to errors in reporting an abnormal patient as normal, or vice versa for borderline cases. This fact has raised the significance of automated measurement systems by reducing the variability and increasing the reliability of cardiac reports among US operators. Furthermore, automation saves a considerable amount of time by improving clinical workflow.   

The problem of automated prediction of clinical measurements, such as segmentation and keypoint localization of anatomical structures, has been approached from different angles, especially within the deep learning literature, where leveraging large size training datasets has led to significant improvements in the accuracy of predicted measurements. Most of the recent methods have used fully convolutional neural networks (FCN) as their main building block to predict pixel-level labels \cite{unet2015u,vnet2016v,avendi2016combined,oktay2017acnn,ngo2017combining,bai2017semi,Bai2018,yao2018weakly}. Similar to numerous works in pose detection literature \cite{pfister2015flowing,tompson2014joint}, in many FCN-based methods, the structure localization problem has been approached by predicting heatmaps corresponding to the regions of interest at some point in the network~\cite{payer2016regressing}. In \cite{payer2016regressing}, a convolutional neural network (CNN) architecture was explored to combine the local appearance of one landmark with the spatial configuration of other landmarks for multiple landmark localization. However, these methods are introduced for problems where data consists of individual frames, rather than temporal sequences. On the contrary, time plays an important role in the calculation of measurements such as EF in cardiac cycles. Therefore, the sole use of these methods may not be sufficient for our problem of interest and other temporally constrained or real-time applications.

Recent studies have made use of spatio-temporal models to overcome limitations of previous models in problems dealing with sequential data, and particularly, echo cine loops \cite{savioli2018automated, dezaki2018cardiac}. In \cite{sofka2017fully}, while a center of the mass layer was introduced and placed on top of an FCN architecture to regress keypoints out of the predicted heatmaps directly, a convolutional long short-term memory (CLSTM) network was also utilized for improving temporal consistency. In the cardiac segmentation domain, many works such as \cite{du2019cardiac} have applied recurrent neural networks to their pipeline. In \cite{li2019recurrent}, multi-scale features are first extracted with pyramid ConvBlocks, and these features are aggregated using hierarchical ConvLSTMs. Other types of studies have fed motion information to their network based on estimating the motion vector between consecutive frames \cite{yan2018left,jafari2018unified,chen2019tan}. Another case of this method is presented by \cite{qin2018joint}, in which similar to our weakly-supervised problem, motion estimation is obtained from an optical flow branch to enforce spatio-temporal smoothness over a weakly supervised segmentation task with sparse labels in the temporal dimension. However, optical flow estimation might contain drastic errors in consecutive frames with large variability, especially in US images where the boundaries are fuzzy, and considerable amounts of noise and artifacts may be present. Therefore, they may not be suitable for a weakly supervised task where the labels are distant in the time domain. Moreover, although most of the mentioned methods take temporal coherence into account, these constraints may not be directly enforced on the model in a desired way \cite{yan2018left,jafari2018unified,savioli2018automated,qin2018joint,li2019recurrent,chen2019tan,du2019cardiac}. In order to overcome these shortcomings, \cite{weitemporal} proposed a method for consistent segmentation of echocardiograms in the time dimension, where only end-diastolic and end-systolic frames have segmentation labels per cycle. This method consists of two co-learning strategies for segmentation and tracking, in which the first strategy estimates shape and motion fields in appearance level, and the second one imposes further temporal consistency in shape level for the previous segmentation predictions. In our method, however, instead of a segmentation task, we perform detection and tracking with reciprocal learning in a landmark detection paradigm in the presence of sparse temporal labels.

\section{Approach}
Our general RDT framework can be found in Figure~\ref{fig:1}. The model can be divided into three parts, the \emph{feature encoder} (blue color), \emph{detection head} (orange color), and \emph{tracking head} (green color). The feature encoder and detection head combined can be viewed as a Unet-like model, for which the general structure is similar to Unet. In the model training phase, the input of the RDT model is an echo sequence starting from the end-diastolic frame and ending at the end-systolic frame. For the detection branch, the input is the whole frame, while for the tracking branch, the inputs are patches from two neighboring frames. The output of the network is two predicted landmark pair locations for LVID.

\subsection{Problem Formulation}
Suppose the frames in the cardiac cine series are represented by $\{I_1, I_2, I_3,..., I_k\}$. For model training, we suppose the end-diastolic frame to be the $1^{st}$ frame, and the end-systolic frame to be the $k^{th}$ frame. The $1^{st}$ and $k^{th}$ frames are with annotation, while the in-between frames are unannotated. The landmark pairs are represented by ${i_t, a_t}$ ($i_t = \{x^i_t, y^i_t\}, a_t = \{x^a_t, y^a_t\}$) corresponding to the landmarks on the inferolateral and anteroseptal walls of LV in the $t^{th}$ frame, respectively. We use $\phi$ to represent the \emph{feature encoder}, and the feature generated for $I_t$ is represented by $\phi_{I_t}$. The $\phi_{I_{t}}$ is solely input to the \emph{detection head} $D$ to get the predicted landmark locations ${i_t^D, a_t^D}$. For \emph{tracking head}, the input is the cropped features of two consecutive frames. One serves as the template frame while the other serves as the search frame. For landmark tracking, the predicted locations start from the $2^{nd}$ frame. After a cycle forward and backward propagation, the predicted location will end at the $1_{st}$ frame.

\subsection{Network Architecture and Losses}
\subsubsection{Shared Feature Encoder}
The feature encoder consists of six 3$\times$3 convolution layers, each followed by a rectified linear unit (ReLU). The third convolution layer is with a stride equal to 2. Since a single feature encoder is sufficient for the tracking head, we share this part of the encoder with both tracking and detection branches. Since the shared encoder is optimized by losses generated from different heads, the encoded feature should be robust since its optimization considers both the spatial information exploited by the detection branch and temporal information explored by the tracking branch.

\subsubsection{Detection Head and Focal Loss}
The detection head combined with the feature encoder together can be viewed as an Unet-like structure, which consists of a contracting path and an expansive path. The contracting path follows the typical architecture of a convolutional network. The beginning of the detection head is another six layers for feature generation. There are two similar downsampling steps to the shared feature encoder. However, we also double the number of feature channels in these two steps. Every step in the expansive path consists of an upsampling of the feature map followed by a 2$\times$2 convolution (“up-convolution”). The first two upsampling layers halve the number of feature channels. We also concatenate the output of each upsampling layer with a correspondingly cropped feature map from the contracting path. Each 3$\times$3 convolutions is followed by a ReLU. As padding is applied, there is no cropping in the whole neural network. For the final two layers used for classification, the first one is a 3$\times$3 convolution layer, and the second is a 1$\times$1 layer, which is used to map each 48-component feature vector to the desired number of landmarks (Here, the number of landmarks is 2). The last layer's output is a two-dimension heatmap, and each location of the heatmap represents the probability of a target landmark.

Focal loss is generated on annotated frames. For each landmark, there is one ground-truth positive location in each dimension of the heatmap (Two landmarks correspond to two dimensions), and all the other locations are negative. For such ground truth, penalizing negative location equally with the positive ones is not appropriate, therefore we apply the focal loss. During training, we reduce the penalty given to negative locations within a radius of the positive location. We empirically set the radius to be 10 pixels. The amount of penalty reduction is given by an unnormalized 2D Gaussian $e^{-(x^2+y^2)/2\delta^2}$, whose center is at the positive location and whose $\sigma$ is 1/3 of the radius. Let $p_{c_{i,j}}$ be the score at location (i, j) for landmark c in the predicted heatmap, and let $y_{c_{i,j}}$ be the ground-truth heatmap augmented with the unnormalized Gaussians. We create a variant of focal loss \cite{lin2017focal}:
\begin{equation}
\scriptsize
{\mathcal{L}_{\det }} = \sum\limits_{c = 1}^2 {\sum\limits_{i = 1}^H {\sum\limits_{j = 1}^W {\left\{ {\begin{array}{*{20}{c}}
{{{(1 - {p_{c_{i,j}}})}^\alpha }\log ({p_{c_{i,j}}})\quad if\quad {y_{c_{i,j}}} = 1}\\
{{{(1 - y{}_{c_{i,j}})}^\beta }{{({p_{c_{i,j}}})}^\alpha }\log (1 - {p_{c_{i,j}}})\quad else,}
\end{array}} \right.} } }     \label{eq:0}
\end{equation}
where $\alpha$ and $\beta$ are the hyperparameters that control the contribution of each point (we empirically set $\alpha$ to 2 and $\beta$ to 4 in all experiments). With the Gaussian distribution encoded in the $y_{c_{i,j}}$, the term $1 - {y_{c_{i,j}}}$ is used for reducing the penalty around the ground truth locations.

\subsubsection{Tracking Head and Cycle Loss:}
For the tracking head, when we get $\phi_{I_{t}}$ and $\phi_{I_{t-1}}$, we first crop the search patches and the template patches both centering at the landmark pairs in the two consecutive frames, respectively. The two template patches for inferolateral/anteroseptal landmarks get concatenated and are represented by $P_{t-1}$, while the two search patches for inferolateral/anteroseptal landmarks get concatenated and are represented by $N_t$.

The input for the tracking branch is the template patch $P_{t-1}$ with size $25\times25$ and the search patch $N_t$ with size $29\times29$, both centering at the landmark patch $i_{t-1}, a_{t-1}$. The size of $P_{t-1}$ and $N_t$ are labeled in Fig. \ref{fig:1}, which are set empirically. We formulate the \emph{tracking head} $T$ as $\delta_{i_t}, \delta_{a_t} = T(\phi_{P_{t}}, \phi_{N_{t+1}})$.

For the tracking head, we first define a convolutional operation between $\phi_{P_{t-1}}$ and $\phi_{N_t}$ in order to compute the affinity (similarity) between each sub-patch of $\phi_{N_t}$ and $\phi_{P_{t-1}}$. To be more specific, $\phi_{P_{t-1}}$ and $\phi_{N_t}$ are combined by using a cross-correlation layer

\begin{equation}
f(\phi_{N_t}, \phi_{P_{t-1}}) = \phi_{P_{t-1}} * \phi_{N_t} .
\end{equation}
Note that the output of this function is a feature map indicating the \emph{affinity score}. For hands-on implementation, it is simple to take $\phi_{P_{t-1}}$ as a kernel matrix to compute dense convolution on $\phi_{N_t}$ within the framework of existing conv-net libraries. The output feature map is followed by another three fully connected layers (represented by m in Eq. \ref{eq:3-1}) to predict the landmark motion. Such regression operation is further formulated as
\begin{equation}
    T(\phi_{P_{t}}, \phi_{N_{t+1}}) = \delta_{i_t}, \delta_{a_t} =m(f(\phi_{N_t}, \phi_{P_{t-1}});\theta_f)  .\label{eq:3-1}
\end{equation}
where $\theta_f$ represents the parameters for the fully connected network. $\delta_{i_t}$ and $\delta_{a_t}$ are both two-dimensional moves (along x-axis and y-axis, respectively). The new landmark location is calculated by adding its previous location to the predicted motion. Such motion prediction is generally similar with optical flow, in which a new three-layer regression is also incorporated. This regression makes the learning process adaptive.

\begin{figure*}
\centering
\includegraphics[width=\linewidth]{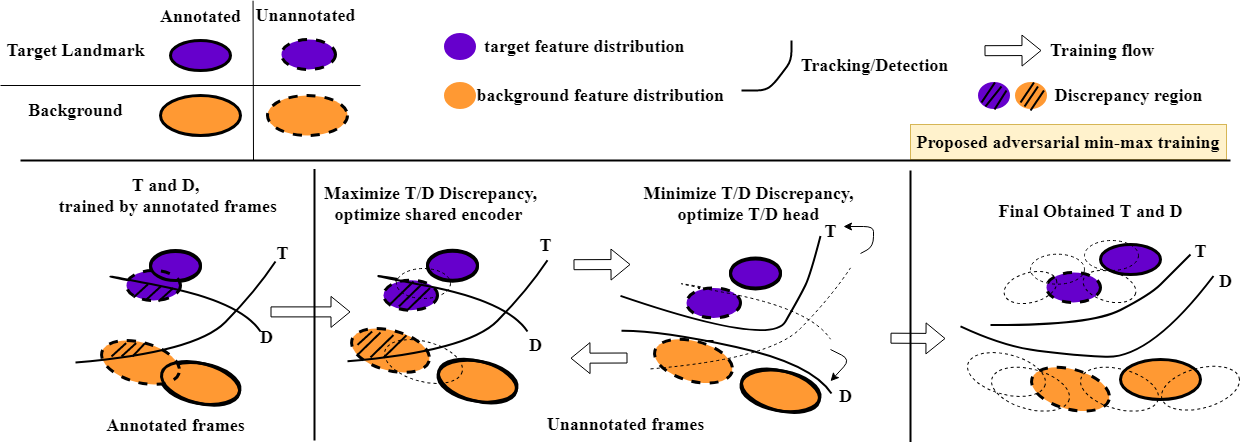}
\caption{Optimization of the proposed reciprocal training.}
\label{fig:3}
\end{figure*}

As the tracking process is only supervised by end-diastolic and end-systolic frames, we introduce the cycle loss and motion loss to supervise the tracking branch. To model the cycle process, we iteratively apply the tracking head $T$ in a forward manner:
\begin{equation}
\begin{array}{l}
{L_t}^* = T({\phi _{{P_{t - 1}}}},{\phi _{{N_t}}}) + {L_{t-1}}^*\\
 = T({\phi _{{P_{t - 1}}}},{\phi _{{N_t}}}) + T({\phi _{{P_{t - 2}}}},{\phi _{{N_{t - 1}}}}) + {L_{t-2}}^*\\
 = T({\phi _{{P_{t - 1}}}},{\phi _{{N_t}}}) + ... T({\phi _{{P_1}}},{\phi _{{N_2}}}) + {L_1}^* , \label{eq:3}
\end{array}
\end{equation}
in which $L_t^*=\{i_t,a_t\}$ represents the predicted location of landmark pairs in $t^{th}$ frame, while $L_1^*=\{i_1,a_1\}$ represents the ground truth location of landmark pairs in the first annotated frame. Here "+" represents the element-wise addition between the location of landmarks in the current frame and motion calculated in Eq.\ref{eq:3-1}. Also, we use the same formulation in backward manner as:
\begin{equation}
\begin{array}{l}
{L_1}^* = T({\phi _{{P_2}}},{\phi _{{N_1}}}) + {L_2}^*\\
 = T({\phi _{{P_2}}},{\phi _{{N_1}}}) + T({\phi _{{P_3}}},{\phi _{{N_2}}}) + {L_3}^*\\
 = T({\phi _{{P_t}}},{\phi _{{N_{t - 1}}}}) + ...T({\phi _{{P_2}}},{\phi _{{N_1}}}) + {L_t}^* . \label{eq:4}
\end{array}
\end{equation}

We use the labeled end-diastolic frame as the beginning frame of the echo cine series, and the end-systolic frame as the end frame. The motion loss is defined by the deviation between the predicted landmark pair locations in the end-systolic frame and their ground truth locations. Suppose the labeled end-systolic frame is the $k^{th}$ frame; after forward propagation, the motion loss $\mathcal{L}$ is defined as
\begin{equation}
\begin{array}{l}
\mathcal{L}_{motion}^k = \mathcal{L}_{1\rightarrow k} = \|{L_k} - {L_k}^*\|^2\\
 = \|{L_k} - (T({\phi _{{P_{k - 1}}}},{\phi _{{N_k}}}) + ...T({\phi _{{P_1}}},{\phi _{{N_2}}}) + {L_1})\|^2 .\label{eq:5}
\end{array}
\end{equation}
The forward propagation is followed by backward propagation that ends at the end-diastolic frame. By combining Eq.~\ref{eq:3} and Eq.~\ref{eq:4}, the current predicted landmark pair location in the diastolic frame $L_1^*$ can actually be represented by its ground truth location $L_1$, and we use the deviation between these two terms to represent the cycle loss as follow:
\begin{equation}
\begin{array}{l}
\mathcal{L}_{cycle}^k = \mathcal{L}_{1\rightarrow k\rightarrow 1} = \|{L_1} - {L_1}^*\|^2\\
 = \|{L_1} - {L_k}^* + {L_k}^* - {L_1}^*\|^2\\
 = \|(T({\phi _{{P_{k - 1}}}},{\phi _{{N_k}}}) + ...T({\phi _{{P_1}}},{\phi _{{N_2}}})) +\\
 (T({\phi _{{P_k}}},{\phi _{{N_{k - 1}}}}) + ...T({\phi _{{P_2}}},{\phi _{{N_1}}}))\|^2.
\end{array}\label{eq:6}
\end{equation}
Finally, the cycle loss can be simplified as
\begin{equation}
\mathcal{L}_{cycle}^k =  - (\mathcal{L}_{motion}^k + \mathcal{L}_{motion}^1).\label{eq:7}
\end{equation}

\subsubsection{Reciprocal Loss for Unannotated Frames:}
The former motion loss, cycle loss, and focal loss are applied for the annotated frames, whereas the reciprocal loss is proposed only for the unannotated frames, which can be viewed as a self-supervision. In the training phase, only the end-diastolic and end-systolic frames are annotated while the in-between frames are unannotated. For these unannotated frames, we can generate both the $i_t^D, a_t^D =  max(D(\phi_{I_t}))$ and the $i_t^T, a_t^T =  T(\phi_{P_{t-1}}, \phi_{N_{t}}) + i_{t-1}^T, a_{t-1}^T$. Although no annotation was assigned, the two predicted landmark pair locations are assumed to be the same. The discrepancy between these two formulates the reciprocal loss. The frame rate for reciprocal loss is set as 3, which means such loss is generated in every three frames. As $D(\phi_{I_t})$ is a heatmap with each location indicating the probability of target location, we define the reciprocal loss similar to the focal loss. We assume $i_t^T$ and $a_t^T$ to be the only positive locations of frame $t$, which is augmented as a 2D Gaussian distribution centering at each positive location. The predicted heatmap from the detection branch is viewed as predicted locations. The formulated reciprocal loss ${\mathcal{L}_{rec}(D, T)}$ is the same as defined in Eq.~\ref{eq:0}.

\section{Optimization}
The basic idea for the proposed RDT model is to create a reciprocal learning between the detection task and the tracking task, as the detection task mainly focuses on the spatial information of a single frame, while the tracking task considers the temporal correlation between consecutive frames. However, the detected landmark pair locations and the tracked landmark pair locations are assumed to be the same. Therefore, we want the two branches to generate a discrepancy to optimize both the feature encoder $\phi$ and the detection/tracking head.

We propose a novel adversarial optimization mechanism. The motivation is for feature augmentation as the number of data is really limited. Trained by the augmented feature, both the detection head D and the tracking head T in Fig. \ref{fig:3} can be more robust. In Fig. \ref{fig:3}, we use blue color to represent the feature distribution of the target landmark pair, and orange color to represent the background. In order to generate a more different distribution of features from unannotated frames, we propose to utilize the disagreement between D and T on the prediction of unannotated frames. We assume D and T can predict the location of annotated frames correctly. Here, we use a key intuition that the feature distribution of unannotated data outside the support of the annotated ones is likely to be predicted differently by D and T. Black lines denote this region as in Fig. \ref{fig:3} (Discrepancy Region). Therefore, if we can measure the disagreement between D and T and train $\phi$ to maximize the disagreement, the encoder will generate more unknown feature distributions outside the support of the annotated ones. The disagreement here is our formerly formulated reciprocal loss ${\mathcal{L}_{rec}(D, T)}$. This goal can be achieved by iterative steps as in Fig. \ref{fig:4}. We first update the feature encoder to maximize the ${\mathcal{L}_{rec}(D, T)}$. Then we freeze this encoder part, and update the D and T to minimize the ${\mathcal{L}_{rec}(D, T)}$, in order to get the uniformed predicted results for the newly generated unknown feature from the feature encoder. Detailed optimization steps are described as follows.

\subsection{Training Steps:}
\begin{figure}
\centering
\includegraphics[width=1.07\linewidth]{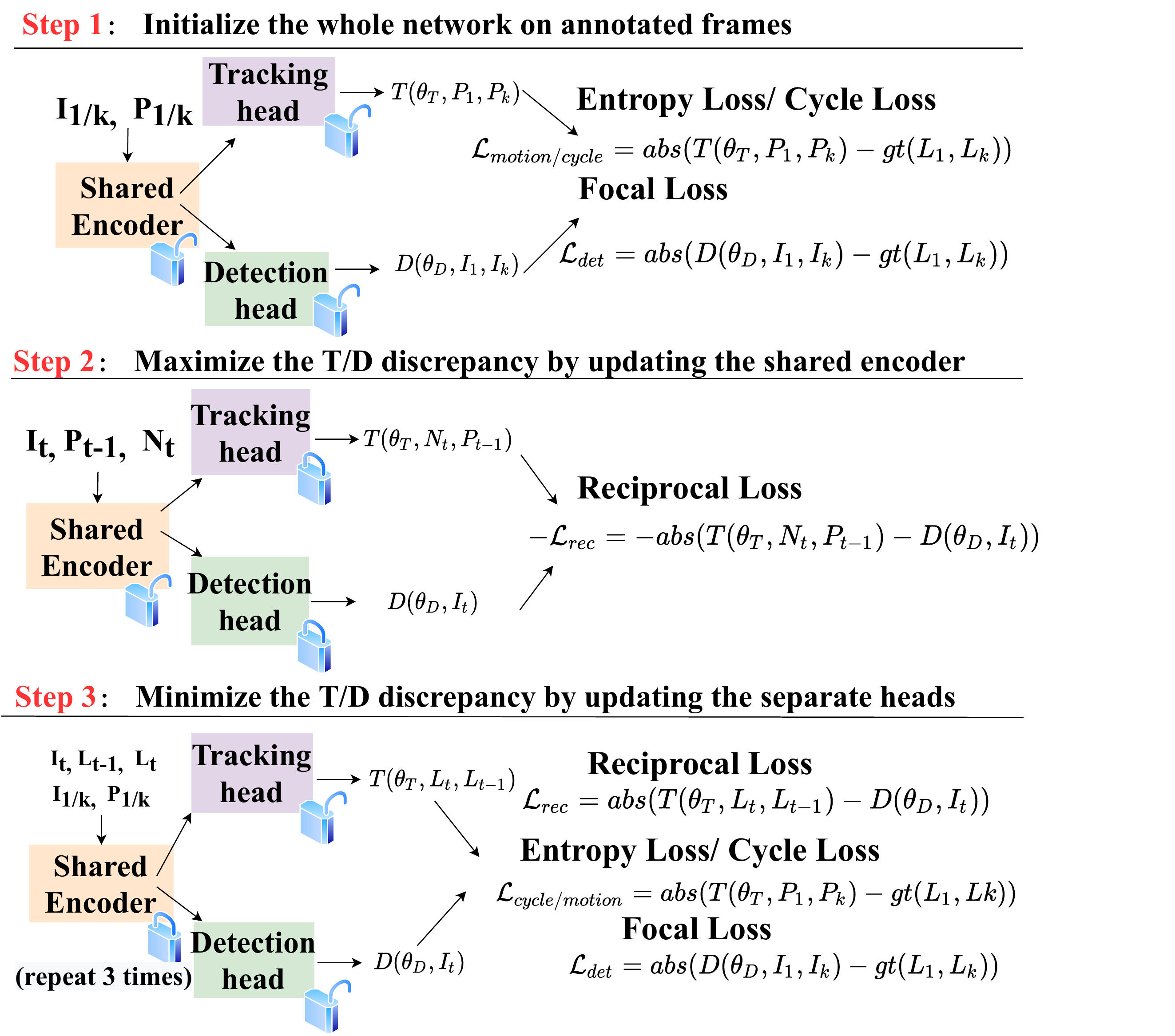}
\caption{Stepwise model training process.}
\label{fig:4}
\end{figure}

We need to train D and T, which take inputs from $\phi$. Both D and T must predict the annotated landmark pair locations correctly. We solve this problem in three steps, as can be found in Fig.~\ref{fig:4}.

\textbf{Step A.} First, we train D, T, and $\phi$ to predict the landmark pairs of annotated frames correctly. We train the networks to minimize three losses applied to annotated frames. The objective is as follows:
\begin{equation}
\mathop {\min }\limits_{\phi ,D,T} ({\mathcal{L}_{\det }} + \mathcal{L}_{motion}^k + \mathcal{L}_{cycle}^k) ;
\end{equation}

\textbf{Step B.} In this step, we train the feature encoder $\phi$ for fixed D and T. By training the encoder to increase the discrepancy, more unknown feature distributions different from the annotated data can be generated. Note that this step only uses the unannotated data. The objective can be formulated as:
\begin{equation}
\mathop {\max }\limits_\phi  ({\mathcal{L}_{rec}}({\rm{D}},T)) ;
\end{equation}

\textbf{Step C.} We train D and T to minimize the discrepancy with a fixed $\phi$. As this step is to get the uniformed and correct detection/tracking results, the step is repeated for three times for the same mini-batch empirically. This setting achieves a trade-off between the encoder and the heads (detection, tracking). This step is applied on both annotated and unannotated frames, to get the best model weights of detection/tracking heads for all the existing features. The objective is as follows:
\begin{equation}
\mathop {\min }\limits_{D,T} ({\mathcal{L}_{\det }} + \mathcal{L}_{motion}^k + \mathcal{L}_{cycle}^k + {\mathcal{L}_{rec}}({\rm{D}},T)) .
\end{equation}

These three steps are repeated until convergence. Weights for different losses are emprically set as 1, in both Step A and Step C. Based on our experience, the order of the three steps is not essential. However, our primary concern is to train D, T, and $\phi$ in an adversarial manner.

\section{Experiments}
\subsection{Dataset and Setup} 
Our echocardiography dataset is collected from our local hospital, following approvals from the Medical Research Ethics Board in coordination with the privacy office. Data were randomly split into mutually exclusive training and testing datasets, where no patient data were shared across the two datasets. The training dataset includes 995 echo cine series with 1990 annotated frames, while the testing dataset includes 224 sequences with 448 annotated frames. Different sequences have a different number of frames ranging from 10s to 100s. The number of frames between end-diastolic and end-systolic phases is different for each cine sample, ranging from 5 to 20 frames.

We run the experiments on our 8x Tesla V100 Server. For the hardware, the CPU is Intel(R) Xeon(R) CPU E5-2698 v4. All comparison methods are trained until convergence. For the proposed method trained by a single GPU, the model converges at 30 epochs, and the running time is 31min/epoch.

\subsection{Quantitative Results} 
EF in PLAX view is estimated based on the distance between inferolateral and anteroseptal landmarks, i.e. LVID. We use the length error (LE) of LVID as well as the location deviation error (LDE) of inferolateral/anteroseptal landmarks (abbreviated as IL/AL) as key errors. LDE is also the most widely used criterion for detection/tracking methods. The comparison is mainly made among the proposed method, the most recently proposed frame by frame detection-based method (Modified U-Net\cite{gilbert2019automated}, CenterNet\cite{zhou2019objects}), and the regular detection+tracking method (Unet+C-Ynet\cite{lin2020cyet}). Unet here is with the same structure as the proposed method. Unet and C-Ynet are trained separately. A general comparison can be found in Table \ref{tab:1}.
\begin{table*}[t]\footnotesize
\begin{center}
\caption{Statistical comparison with the state-of-the-art methods. Errors ('cm') for different sequences are sorted in ascending order. Evaluation criteria are the Length Error (LE) and the Location Deviation Error (LDE) of Inferolateral/Anteroseptal Landmarks (IL/AL)} \label{tab:1}
\begin{tabular}{c|c|c c c c c c c c}
  \hline

  \textbf{Method} & \textbf{Frame} & \textbf{Criterion(cm)}  & \textbf{Mean$\pm$ std} & \textbf{min} & $25\%$ & \textbf{Median} & $75\%$ & $90\%$ & \textbf{max}
  \\
  \hline
   &  & LDE of AL  & \textbf{1.28 $\pm$ 1.43} & \textbf{0.01} & \textbf{0.51} & \textbf{0.96} & \textbf{1.60} & \textbf{2.26} & \textbf{11.71}\\
  Proposed RDT & end-diastolic & LDE of IL & \textbf{1.16$\pm$1.27} & 0.06 & \textbf{0.40} & \textbf{0.88} & \textbf{1.48} & \textbf{2.15} & \textbf{10.67} \\
  & & LE of LVID & \textbf{0.81$\pm$1.04} & \textbf{0.00} & \textbf{0.25} & \textbf{0.51} & \textbf{1.00} & \textbf{1.63} & \textbf{8.07}\\
    \hline
   CenterNet &  & LDE of AL  & 1.79 $\pm$ 1.96 & 0.07 & 0.72 & 1.21 & 2.04 & 3.81 & 13.62\\
  \cite{zhou2019objects} & end-diastolic & LDE of IL & 1.71$\pm$1.82 & 0.09 & 0.61 & 1.34 & 2.49 & 3.13 & 12.60 \\
  & & LE of LVID & 1.22$\pm$1.94 & 0.03 & 0.44 & 1.15 & 1.81 & 2.24 & 10.33\\
    \hline

   Unet+C-Ynet &  & LDE of AL & 2.29$\pm$3.02 & 0.05 & 0.68 & 1.41 & 2.35 & 5.21 & 19.01 \\
   \cite{lin2020cyet}  & end-diastolic & LDE of IL & 3.72$\pm$4.05 & 0.07 & 0.78 & 1.91 & 5.26 & 10.89 & 18.81 \\
 & & LE of LVID & 2.39$\pm$2.61 & \textbf{0.00} & 0.64 & 1.38 & 3.28 & 5.77 & 12.16\\
    \hline

  Modified U-Net&  & LDE of AL & 5.15$\pm$4.86 & 0.10 & 1.27 & 2.99 & 8.18 & 12.72 & 19.76 \\
  \cite{gilbert2019automated} & end-diastolic & LDE of IL & 5.36$\pm$4.74 & \textbf{0.03} & 1.01 & 4.13 & 8.86 & 12.31 & 17.22\\
  & & LE of LVID & 3.40 $\pm$ 3.02 & 0.02 & 0.97 & 2.49 & 5.07 & 7.63 & 15.17\\
  \hline
  \hline

   &  & LDE of AL & \textbf{1.44 $\pm$ 1.30} & \textbf{0.06} & \textbf{0.66} & \textbf{1.16} & \textbf{1.75} & \textbf{2.67} & \textbf{10.37} \\
  Proposed RDT & end-systolic & LDE of IL & \textbf{1.13$\pm$1.22} & 0.06 & \textbf{0.51} & \textbf{0.90} & \textbf{1.25} & \textbf{1.89} & \textbf{10.10} \\
  & & LE of LVID & \textbf{1.09$\pm$0.95} & \textbf{0.00} & \textbf{0.37} & \textbf{0.90} & \textbf{1.51} & \textbf{2.43} & \textbf{5.81}\\
  \hline
  CenterNet &  & LDE of AL & 1.90 $\pm$ 1.64 & 0.09 & 0.98 & 1.73 & 2.98 & 3.75 & 13.57 \\
  \cite{zhou2019objects} & end-systolic & LDE of IL & 2.03$\pm$2.21 & 0.12 & 0.92 & 1.98 & 3.68 & 4.42 & 14.54 \\
  & & LE of LVID & 1.83$\pm$1.48 & 0.06 & 0.95 & 1.78 & 2.93 & 4.31 & 9.25\\
  \hline
   Unet+C-Ynet &  & LDE of AL & 2.78$\pm$2.87 & 0.14 & 0.98 & 1.82 & 3.29 & 5.85 & 19.8 \\
   \cite{lin2020cyet}  & end-systolic & LDE of IL & 3.42$\pm$3.80 & 0.06 & 0.78 & 1.74 & 4.71 & 9.73 & 17.33 \\
    & & LE of LVID & 2.45$\pm$2.61 & 0.00 & 0.73 & 1.41 & 3.02 & 5.14 & 11.51\\
    \hline

  Modified U-Net&  & LDE of AL & 5.05$\pm$4.34 & 0.16 & 1.42 & 2.90 & 8.47 & 12.04 & 16.79\\
  \cite{gilbert2019automated} & end-systolic & LDE of IL & 5.72$\pm$4.59 & \textbf{0.03} & 1.70 & 4.65 & 9.26 & 12.24 & 17.91\\
  & & LE of LVID & 3.87$\pm$3.21 & 0.03 & 1.64 & 3.02 & 5.18 & 8.39 & 19.38\\
  \hline
\end{tabular}
\end{center}
\end{table*}
Comparison with state-of-the-art methods is reported in Table~\ref{tab:1}. Our results verify that our detection on the end-diastolic frame performs best over compared methods. Results also demonstrate that errors in end-systolic and end-diastolic frames are of the same range, suggesting that the tracking error is not accumulative over in-between unannotated frames.

\begin{figure}
\centering
\includegraphics[width=\linewidth]{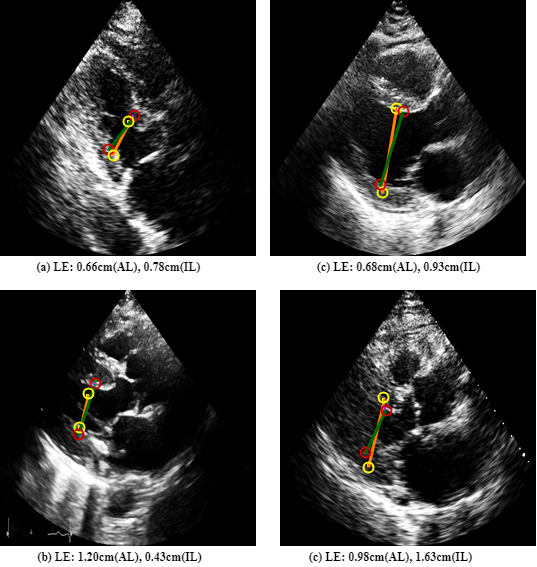}
\caption{Four examples of frames with median LDE. The predicted LVID is the orange color line with landmarks in yellow color, while the ground truth LVID is the green color line with landmarks in red color.}
\label{fig:5}
\end{figure}

\subsection{Qualitative Results with Visualized Examples}
 Fig. \ref{fig:5} shows four examples with the location error around the median. Here the Location Deviation Error (LDE) is the average location error of Inferolateral/Anteroseptal Landmarks (AL and IL), as there are no cases in our test data for which the AL and IL are both at the median. For the end-systolic frame, the average LDE is 0.95$\pm$0.68~cm for mean $\pm$ std, and 0.85~cm for the median. For the end-diastolic frame, the average LDE is with 0.91$\pm$0.66~cm for mean $\pm$ std, and 0.82~cm for the median.

\subsection{Ablation Study}
In our ablation study, we verify the effectiveness of the adversarial training (Ad-T), as well as the reciprocal loss (Rec-L). Without the reciprocal loss, the structural information of in-between unannotated frames is ignored. As Ad-T is based on Rec-L, without Rec-L the Ad-T cannot be achieved. A detailed comparison can be found in Table \ref{tab:2}.

\begin{table}[t]\footnotesize
\begin{center}
\caption{Ablation study for Ad-T and Rec-L.} \label{tab:2}
\begin{tabular}{c c c c c c}
  \hline

  \textbf{Frame} & \textbf{Criterion(cm)} & \textbf{Ad-T} & \textbf{Rec-L} & \textbf{mean} & \textbf{median}
  \\
  \hline
   & LDE-AL & $\color{blue}{\times}$ & $\color{blue}{\times}$& 3.22 & 4.50 \\
 & LDE-IL &$\color{blue}{\times}$ & $\color{blue}{\times}$& 5.02 & 6.74 \\
   & LE & $\color{blue}{\times}$ & $\color{blue}{\times}$& 2.65 & 3.53 \\

   & LDE-AL & $\color{blue}{\times}$ & $\color{red}{\checkmark}$& 1.70 & 1.95 \\
   ED   & LDE-IL & $\color{blue}{\times}$ & $\color{red}{\checkmark}$& 1.76 & 2.02 \\
   & LE & $\color{blue}{\times}$ & $\color{red}{\checkmark}$& 1.00 & 1.04 \\

   & LDE-AL & $\color{red}{\checkmark}$& $\color{red}{\checkmark}$& \textbf{1.28} & \textbf{0.96} \\
   & LDE-IL &$\color{red}{\checkmark}$ & $\color{red}{\checkmark}$& \textbf{1.16} & \textbf{0.88} \\
   & LE & $\color{red}{\checkmark}$&$\color{red}{\checkmark}$& \textbf{0.81} & \textbf{0.51} \\
  \hline
   & LDE-AL & $\color{blue}{\times}$ & $\color{blue}{\times}$& 3.17 & 3.85 \\
   & LDE-IL & $\color{blue}{\times}$ & $\color{blue}{\times}$& 4.79 & 6.94 \\
   & LE & $\color{blue}{\times}$ & $\color{blue}{\times}$& 2.36 & 3.47 \\

   & LDE-AL & $\color{blue}{\times}$ & $\color{red}{\checkmark}$& 1.76 & 1.92 \\
   ES & LDE-IL &$\color{blue}{\times}$ & $\color{red}{\checkmark}$& 1.88 & 1.94 \\
   & LE & $\color{blue}{\times}$ & $\color{red}{\checkmark}$& 1.41 & 1.54 \\

   & LDE-AL & $\color{red}{\checkmark}$ & $\color{red}{\checkmark}$& \textbf{1.44} & \textbf{1.75} \\
   & LDE-IL &$\color{red}{\checkmark}$ & $\color{red}{\checkmark}$& \textbf{1.13} & \textbf{1.25} \\
   & LE & $\color{red}{\checkmark}$& $\color{red}{\checkmark}$& \textbf{1.09} & \textbf{0.90} \\
   \hline
\end{tabular}
\end{center}
\end{table}

Table \ref{tab:2} shows that the reciprocal loss substantially improves the framework. By adding the reciprocal loss, the errors decrease around 2~cm for all different criteria. The results again improve a lot when the model is trained with our proposed Ad-T method.

\subsection{Model Extension}
We also test our proposed model's extension ability, in which we try only to use \textbf{one frame} (end-diastolic) in each sequence for model training. Such training would start from the first annotated frame, and then track in a cycle way. The motion loss and the focal loss in the last frame are not available in such a model. The model is mainly trained by the reciprocal loss from the unannotated frames and the focal loss as well as the cycle loss in the annotated frame (i.e., the end-diastolic frame). Detailed results are reported in Table \ref{tab:3}. We simply use the medium value of two LDEs (AL and IL) to represent the LDE.
\begin{table}[t]\footnotesize
\begin{center}
\caption{Statistics analysis for model trained by one frame only.} \label{tab:3}
\begin{tabular}{c|c|c c c c c c c c}
  \hline

 \textbf{Frame} & \textbf{Criterion(cm)}  & \textbf{Mean$\pm$ std} & \textbf{min} & \textbf{Medium}
  \\
  \hline
   ED  & LDE & 1.59$\pm$1.85 & 0.04 & 0.95  \\
    & LE &  1.04$\pm$1.20 & 0.02 & 0.68 \\
    \hline
   ES & LDE & 1.76$\pm$1.49 & 0.10 & 1.34 \\
    & LE & 1.77$\pm$1.39 & 0.01 & 1.49\\
  \hline
\end{tabular}
\end{center}
\end{table}

We can find even with only one frame annotated, the proposed model can get satisfying results, when compared with the state-of-the-art. However, results on end-systolic are much worse than on end-diastolic, which means the second annotated frame affects the tracking branch a lot.

\begin{table}[t]\footnotesize
\begin{center}
\caption{Analysis for the annotation sparsity.} \label{tab:4}
\begin{tabular}{c c c c c }
  \hline
\textbf{Annotation rate} & 5-8  & 8-12 & 12-16 & 16-20
  \\
\hline
Average LDE (cm)/ sequence & 0.23 & 0.26 & 0.24 & 0.27\\
\hline
Average LDE (cm) /frame & 0.031 & 0.025  & 0.021 & 0.018\\
\hline
\end{tabular}
\end{center}
\end{table}

\begin{table}[t]\footnotesize
\begin{center}
\caption{Analysis for the sparsity of reciprocal loss.} \label{tab:5}
\begin{tabular}{c c c c c }
  \hline

 \textbf{Loss rate} & 2  & 3 & 4 & 5
  \\
  \hline
   Average LDE (cm)/ sequence & 0.46 & \textbf{0.25} & 0.29 & 0.38\\
  \hline
\end{tabular}
\end{center}
\end{table}
\subsection{Annotation Sparsity Analysis}
\textbf{Sparsity of annotation:} As the number of in-between unannotated frames is random ranging from 5 to 20, such number may influence the tracking branch, while the detection branch may not be affected. Therefore, to analyze the influence of annotation sparsity on tracking, we just start from the ground truth location of the first frame (end-diastolic) to do the tracking. The whole model does not change. We get the predicted location in the second annotated frame (end-systolic), and use the location deviation errors (LE) on this frame for different sequences with different annotation sparsity for evaluation. Results can be found in Table~\ref{tab:4}.

We observe that the proposed method is not affected by the annotation sparsity. The Average LDEs for different sequences are generally the same around 0.25 cm. The average LDE/frame is the Average LDE divided by the in-between frame number. As the reciprocal loss is generated every three frames, whenever a large error is generated from the tracking branch, the discrepancy between detected and tracked location will also be large, which brings a significant reciprocal loss. Such loss overcomes the problem brought by large annotation sparsity.

\textbf{Sparsity of reciprocal loss:} The frequency for applying reciprocal loss for in-between unannotated frames is also important. A comparison can be found in Table~\ref{tab:5}. We can find that only when the reciprocal loss is applied every three frames, the results are best. Therefore, we empirically set such rate as 3. The reciprocal loss should not be applied too densely or with a large sparsity.

\subsection{Failed Cases Analysis}
\begin{figure}
\centering
\includegraphics[width=0.6\linewidth]{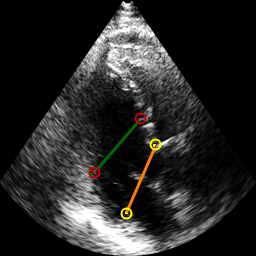}
\caption{An example of a discrepant case. This PLAX view is suboptimal and has been imaged at a low imaging window on the chest resulting altered axis of the LV. The ground truth LVID label (shown in green color), used clinically, has been placed in an atypical position based on operator judgment (closer to the apex) to account for the altered geometry. The predicted LVID is the orange color line with landmarks in yellow color. It should be noted that despite relatively large LDE error, both measurements are likely clinical acceptable, as the distance between AL and IL, rather than their absolute image coordinates, is the main metric used to measure EF.}
\label{fig:6}
\end{figure}

There are still a few failure cases in our current result. An example is shown in Fig.~\ref{fig:6}. The result is with the maximum LDE (LDE of AL: 5.02~cm, LDE of IL: 8.07~cm, LD: 0.48cm). We hypothesize that the reason for such failure is as follows: During the image acquisition, the operator appears to have zoomed the ultrasound image on the LV. Hence, no other cardiac chamber is clearly visible and the appearance of the image is substantially different from a typical PLAX image. A much larger training data set will be required to avoid failure in such cases.

If we set 2~cm error for the average LDE (average of IL and AL) as the critical point for failure, for the end-systolic frame the failure percentage is $6.1\%$, while for the end-diastolic frame the percentage is $3.7\%$. These are promising results, compared with the results in \cite{gilbert2019automated} whose failure is $6.7\%$. We note that the model in \cite{gilbert2019automated} is trained by densely annotated sequences, instead of sparsely annotated sequences as in our method.

\subsection{Ejection Fraction Error Analysis}
We also analyze the proposed method from medical perspective. We calculate the Ejection Fraction error on the testing dataset. Ejection fraction (EF) is a measurement, expressed as a percentage, of how much blood the left ventricle pumps out with each contraction. The ejection fraction represents the percent of the total amount of blood being pushed out with each heartbeat in the left ventricle. The Ejection Fraction (EF) is formulated as
\begin{equation}
    EF = 100 \times (ED_{vol}-ES_{vol})/ED_{vol},
\end{equation}

in which $ED_{vol}$ and $ES_{vol}$ are End-diastolic volume and End-systolic volume respectively, which are formulated by Teichholz formula as below
\begin{equation}
    ED_{vol} = 7\times EDD /(2·4+EDD),
\end{equation}
\begin{equation}
    ES_{vol} = 7\times ESD /(2·4+ESD).
\end{equation}

\begin{figure}
\centering
\includegraphics[width=0.9\linewidth]{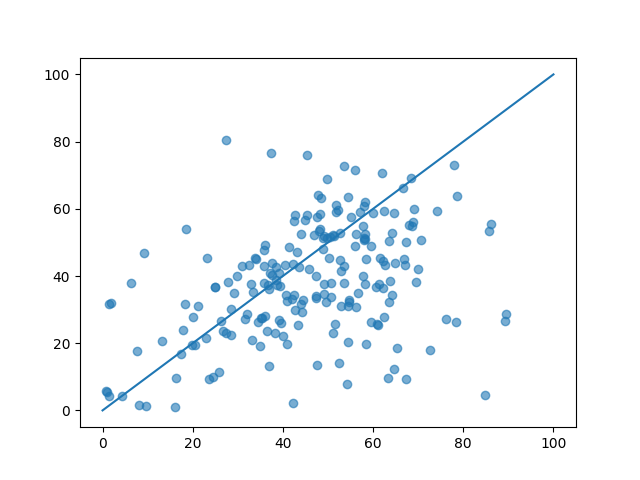}
\caption{The EF scatter plot for the proposed method.}
\label{fig:7}
\end{figure}

Here EDD means the length of LV in the end-diastolic frame and the end-systolic frame respectively. The EF error is the difference between the predicted EF and ground truth EF. The calculated result can be found in Table \ref{tab:1}. Also, we draw the EF scatter as can be found in Fig. \ref{fig:7}.
\begin{table}[t]\footnotesize
\begin{center}
\caption{Statistical result of EF error for the proposed method.} \label{tab:1}
\begin{tabular}{c c c c c}
  \hline
  Results &\textbf{Mean$\pm$ std} & \textbf{min} &\textbf{Median} & \textbf{90\%}
  \\
  \hline
   EF prediction & 37.08 $\pm$17.39 & 0.59 & 37.25 & 63.75\\
  \hline
   EF error & 19.00 $\pm$26.25 & 0.02  & 12.28 & 39.21 \\
  \hline
\end{tabular}
\end{center}
\end{table}

\section{Conclusion}
In this paper, we proposed a novel reciprocal landmark detection and tracking model. The model is designed to tackle the data and annotation scarcity problem for ultrasound sequences. The model achieves reliable landmark detection and tracking with only around 2,000 annotated frames (995 sequences) for training. For each sequence, only two key frames are annotated. The model is optimized by a novel adversarial training way, which can better exploit the training data's limited information. The comparison with state-of-the-art and analysis of results verify the effectiveness of our proposed method.

{\small
\bibliographystyle{IEEEtran}
\bibliography{egbib}
}

\end{document}